# UNDERSTANDING RETAIL PRODUCTIVITY BY SIMULATING MANAGEMENT PRACTICES

Peer-Olaf Siebers[1], Uwe Aickelin[1], Helen Celia[2], Chris W. Clegg[2]

[1]University of Nottingham, School of Computer Science & IT (ASAP)
Nottingham, NG8 1BB, UK
[2]University of Leeds, Centre for Organisational Strategy, Learning & Change, LUBS
Leeds, LS2 9JT, UK

*pos@cs.nott.ac.uk (Peer-Olaf Siebers)*

**Abstract**

Intelligent agents offer a new and exciting way of understanding the world of work. In this paper we apply agent-based modeling and simulation to investigate a set of problems in a retail context. Specifically, we are working to understand the relationship between human resource management practices and retail productivity. Despite the fact we are working within a relatively novel and complex domain, it is clear that intelligent agents could offer potential for fostering sustainable organizational capabilities in the future. Our research so far has led us to conduct case study work with a top ten UK retailer, collecting data in four departments in two stores. Based on our case study data we have built and tested a first version of a department store simulator. In this paper we will report on the current development of our simulator which includes new features concerning more realistic data on the pattern of footfall during the day and the week, a more differentiated view of customers, and the evolution of customers over time. This allows us to investigate more complex scenarios and to analyze the impact of various management practices.

**Presenting Author's biography**

Peer-Olaf Siebers is a Research Fellow in the School of Computer Science & IT at the University of Nottingham. His main research interest is the application of computer simulation to study human oriented complex adaptive systems. Complementary fields of interest include distributed artificial intelligence, biologically inspired computing, game character behavior modeling, and agent-based robotics. His webpage can be found via <www.cs.nott.ac.uk/~pos>.

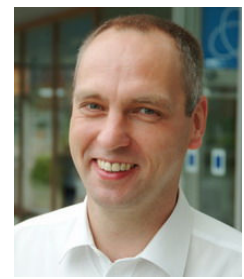

# 1 Introduction

The retail sector has been identified as one of the biggest contributors to the productivity gap that persists between the UK, Europe and the USA [1]. It is well documented that measures of UK retail productivity rank lower than those of countries with comparably developed economies [2]. Intuitively, it seems likely that management practices are linked to a company's productivity and performance.

Significant research has been done to investigate the productivity gap and identify problems involved in estimating the size of the gap; for example the comparability of productivity indices [3], historical influences [4], general measurement issues [5], and varying sectoral contributions [6]. Best practice guidelines have been developed and published, but there remains considerable inconsistency and uncertainty regarding how these are implemented and manifested at the level of the work place. Indeed, a recent report on UK productivity asserted that, "... the key to productivity remains what happens inside the firm and this is something of a 'black box'," [7]. Siebers and colleagues [8] conducted a comprehensive literature review of this research area to assess linkages between management practices and firm-level productivity. The authors concluded that management practices are multidimensional constructs that generally do not demonstrate a straightforward relationship with productivity variables. Empirical evidence affirms that management practices must be context specific to be effective, and in turn productivity indices must also reflect a particular organization's activities on a local level to be a valid indicator of performance.

Currently there is no reliable and valid way to delineate the effects of management practices from other socially embedded factors. Most Operations Research methods can be applied as analytical tools once management practices have been implemented, however they are not very useful at revealing system-level effects prior to the introduction of specific management practices. This is most restricting when the focal interest is the development of the system over time, as happens in the real world. This contrasts with more traditional techniques, which allow us to identify the state of the system at a certain point in time.

The overall aim of our project is to understand and predict the impact of different management practices on retail store productivity. To achieve this aim we have adopted a case study approach using applied research methods to collect both qualitative and quantitative data. In summary, we have worked with a major retail organization to conduct four weeks' of informal participant observations in four departments across two retail stores, forty semi-structured interviews with employees including a 63-item questionnaire on the effectiveness of retail management practices, and drawn upon a variety of established information sources internal to the company. Using this data, we have been applying agent-based modeling and simulation to try to devise a functional representation of the case study departments.

In Section 2, 3 and 4 we give an overview of the research we have conducted to develop a first version of a department store simulator with an infinite customer population. Each customer enters the department only once, however this currently prevents us from investigating long term effects that certain management practices might have. In Section 5 we introduce the second version of the simulator. This simulator is based on the original design, but has been improved to incorporate the specification of a finite population of customers and therefore permits the investigation of long term effects. We have also added a more differentiated view of the customers and other features that are present in the real system and that we believe are important to more accurately model the operation of a department. Finally, Section 6 concludes with a summary and an outlook of further developments planned for our simulator.

# 2 An assessment of alternative modeling techniques

Operations Research is applied to problems concerning the conduct and co-ordination of the operations within an organization [9]. An Operations Research study usually involves the development of a scientific model that attempts to abstract the essence of the real problem. When investigating the behavior of complex systems the choice of an appropriate modeling technique is very important. To inform the choice of technique for the simulator, the relevant literature spanning the fields of Economics, Social Science, Psychology, Retail, Marketing, Operations Research, Artificial Intelligence, and Computer Science was reviewed. Within these fields a wide variety of approaches is used which can be classified into three main categories: analytical approaches, heuristic approaches, and simulation. In many cases we found that combinations of these were used within a single model (e.g. [10,11]). From these approaches we were able to identify simulation as best fitting our needs.

Simulation introduces the possibility of a new way of thinking about social and economic processes, based on ideas about the emergence of complex behavior from relatively simple activities [12]. Simulation allows clarification of a theory and investigation of its implications. While analytical models typically aim to explain correlations between variables measured at one single point in time, simulation models are concerned with the development of a system over time [13].

Operations Research usually employs three different types of simulation modeling to help understand the behavior of organizational systems, each of which has its distinct application area: Discrete Event Simulation (DES), System Dynamics (SD) and Agent Based Simulation (ABS). DES models a system as a set of entities being processed and evolving over time according to the availability of resources and the triggering of events. The simulator maintains an ordered queue of events. DES is widely used for decision support in manufacturing and service industries. SD takes a top down approach by modeling system changes over time. The analyst has to identify the key state variables that define the behavior of the system and these are then related to each other through coupled, differential equations. SD is applied where individuals within the system do not have to be highly differentiated and knowledge on the aggregate level is available, for example modeling population, ecological and economic systems. In an ABS model the researcher explicitly describes the decision processes of simulated actors at the micro-level. Structures emerge at the macro level as a result of the actions of the agents, and their interactions with other agents and the environment [14].

Although computer simulation has been used widely since the 1960s, ABS only became popular in the early 1990s [15]. ABS is well suited to modeling systems with heterogeneous, autonomous and pro-active actors, and can be applied in principle to any human-centered system. ABS is the best option for situations in which individual variability between the agents cannot be neglected. Such models could not be built using SD, because variability between individuals must be defined on the micro level. ABS supports understanding of how the dynamics of real systems arise from the characteristics of individuals and their environment. It allows modeling of a heterogeneous population where each agent can have personal motivations and incentives, and to represent groups and group interactions. These attributes are usually not modeled in DES models, where it is common practice to model people as deterministic resources ignoring their performance variation and their pro-active behaviors. With these simplifications it is not possible to make accurate predictions about the system performance [16]. ABS models take both of these characteristics into account. Each agent's behavior is defined by its own set of attribute values. These variable attributes represent variation in each individual's behavior and the simulation design is decentralised (i.e. the agents are pro-active). ABS is suited to a system driven by interactions between its constituent entities, and can reveal what appears to be complex emergent behavior at the system level even when the agents involved exhibit fairly simple behaviors on a micro-level.

ABS is still a relatively new simulation technique and its principal application has been in academic research. With the availability of more sophisticated modeling tools, things are starting to change [17]. ABS is extensively used by the game and film industry to develop realistic simulations of individual characters and societies. It is used in computer games, for example The SIMS™ [18], or in films when diverse heterogeneous characters animations are required, for example the Orcs in Lord of the Rings™ [19].

Nevertheless, there are some disadvantages associated with the use of ABS. It has a higher level of complexity compared to other simulation techniques, most notably in that all of the interactions between agents and between the agent and the environment have to be defined. Therefore ABS tends to have higher computational requirements.

## 3 Conceptual model design and data collection

Based on the results of our assessment of alternative modeling techniques we started to design conceptual models of the system to be investigated and the actors within the system using the agent paradigm. A conceptual model of the simulator is presented in Figure 1. We have identified three different types of agents: customers, sales staff and managers, each of them having a different set of relevant parameters and we have defined global parameters which can influence any aspect of the system. With regards to the system outputs we always hope to find some unforeseeable, emergent behavior on a macro level. A visual representation of the simulated system and its actors facilitates the monitoring and better understanding of the interactions of entities within the system. Coupled with the standard DES performance measures (e.g. staff utilization, number of sales), we strive to identify bottlenecks and to optimize the modeled system.

The conceptual design of our agents is presented within state charts. State charts show the different states an entity can be in and define the events that cause a transition from one state to another. This is exactly the information we need in order to represent our agents at a later stage within the simulation environment. Furthermore, this form of graphical representation helps with validation of the agent design because it is relatively easy to understand without specialized knowledge.

The art of modeling is simplification and abstraction [20]. A model is always a restricted copy of the real world, and one has to identify the most important components of a system to build effective models. In our case, the important system components take the form of the most important behaviors of an actor and the triggers that initiate a change from one behavior to another. We have developed state charts for all the relevant actors in our department store model. Figure

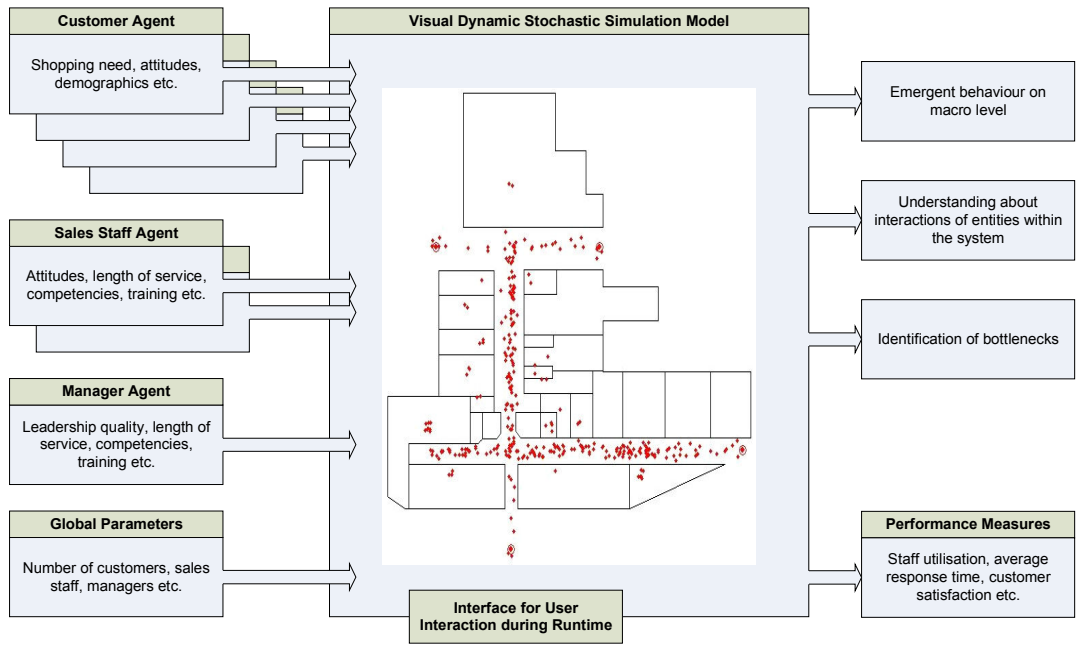

Fig. 1 Conceptual model of the simulator

2 shows as an example the state charts for a customer agent.

Often agents are based on analytical models or heuristics and, in the absence of adequate empirical data, theoretical models are employed. However, we use frequency distributions for state change delays and probability distributions for decision making processes because statistical distributions are the best way in which we can represent the numerical data we have gathered during our case study work. In this way a population is created with individual differences between agents, mirroring the variability of attitudes and behaviors of their real human counterparts.

We collected data in the Audio & Television (A&TV) and the Womenswear (WW) departments of two branches of a leading UK department store. The frequency distributions are modeled as triangular distributions defining the time that an event lasts, using the minimum, mode, and maximum duration and these figures are based on our own observations and expert estimates in the absence of objective numerical data. The probability distributions are partly based on company data (e.g., the rate at which each shopping visit results in a purchase, hereafter referred to as 'conversion rate') and partly on informed estimates (e.g., the patience of customers before they would leave a queue). We also gathered some company data about work team numbers and work team composition, varying opening hours and peak times, along with other operational details.

## 4 ManPraSim v1: The initial simulator

### 4.1 Model implementation (ManPraSim v1)

Our Management Practices Simulator (ManPraSim) has been implemented in AnyLogic™ which is a Java™ based multi-paradigm simulation software

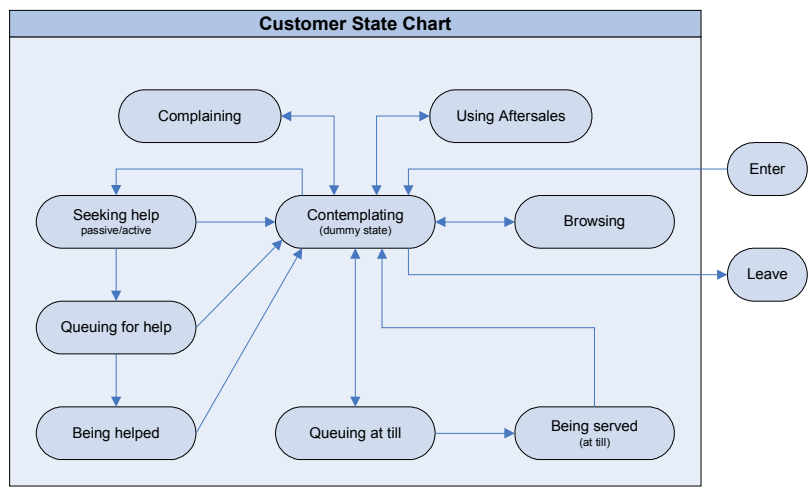

Fig. 2 Conceptual model of the customer agent (transition rules are omitted to keep the chart comprehensible)

[21]. During the implementation we have applied the knowledge, experience and data accumulated through our case study work. The simulator can represent the following actors: customers, service staff (including cashiers, selling staff of two different training levels) and managers. Figure 3 shows a screenshot of the current customer and staff agent logic as it has been implemented in AnyLogic™. Boxes show customer states, arrows transitions, arrows with a dot on top entry points, circles with a B inside branches, circles with a dot inside termination points, and numbers satisfaction weights.

Currently there are two different types of customers implemented: customers who want to buy something and customers who require a refund. If a refund is granted, a customer can change his or her goal to making a new purchase. The customer agent template consists of three main blocks which all use a very similar logic. In each block, in the first instance, a customer will try to obtain service directly and if he or she cannot obtain it (no suitable staff member available) they will have to queue. He or she will then either be served as soon as the right staff member becomes available, or will leave the queue if he or she does not want to wait any longer (an autonomous decision). A complex queuing system has been implemented to support different queuing rules. In comparison to the customer agent template, the staff agent template is relatively simple. Whenever a customer requests a service and the staff member is available and has the right level of expertise for the task requested, the staff member commences this activity until the customer releases the staff member. Whereas the customer is the active component of the simulation model, the staff member is currently passive, simply reacting to requests from the customer.

A service level index is introduced as a novel performance measure using the satisfaction weights mentioned earlier. Historically customer satisfaction has been defined and measured in terms of customer satisfaction with a purchased product [22]. The development of more sophisticated measures has moved on to incorporate customers' evaluations of the overall relationship with the retail organization, and a key part of this is the service interaction. Indeed, empirical evidence suggests that quality is more important for customer satisfaction than price or value-for-money [23], and extensive anecdotal evidence indicates that customer-staff service interactions are an important determinant of quality as perceived by the customer.

The index allows customer service satisfaction to be recorded throughout the simulated lifetime. The idea is that certain situations might have a bigger impact on customer satisfaction than others, and therefore weights can be assigned to events to account for this. Applied in conjunction with an ABS approach, we expect to observe interactions with individual customer differences, variations which have been empirically linked to differences in customer satisfaction (e.g. [24]). This helps the analyst to find out to what extent customers underwent a positive or negative shopping experience. It also allows the analyst to put emphasis on different operational aspects and try out the impact of different strategies.

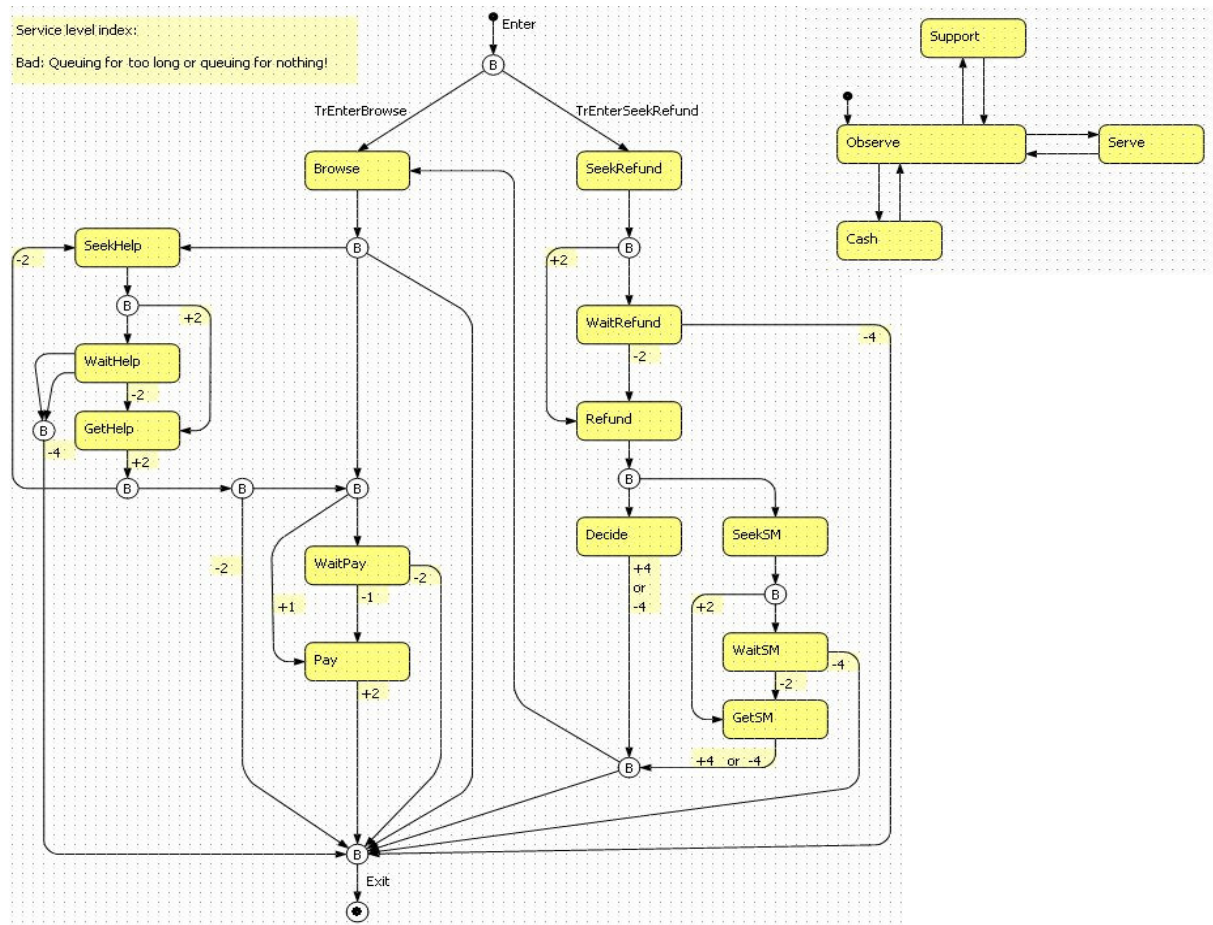

Fig. 3 Customer (left) and staff (right) agent logic implementation in AnyLogic™

### 4.2 Experiments (ManPraSim v1)

In order to test the operation of our simulator and ascertain face validity we have completed several experiments. Here we present two of them in brief. Full details can be found in [25].

Each experiment is grounded in both theory and practice. We have built the simulation model to allow us to observe and evaluate the impact of training and empowerment practices in terms of multiple outcome variables, specifically the volume of sales transactions and various customer satisfaction indices. We use different setup parameters for the departments to reflect real differences between them, for example different customer arrival rates and different service times. All experiments hold the overall number of staffing resources constant at ten staff and the simulation lifespan is always ten weeks.

In the first experiment we varied the staffing arrangement within the pool of ten staff. In each department the staff are allocated to either selling or cashier duties. We predicted a curvilinear relationship between the number of cashiers and each of the outcome variables, and these hypotheses were largely confirmed. We expected this type of relationship because there are limiting factors for the more extreme experimental conditions. Very small numbers of cashiers available to process purchase transactions detrimentally impact on the volume of customer transactions, and at the other end of the scale very small numbers of selling staff restrict scope for customer advice and negatively influence customer perceptions of satisfaction. We also predicted that the peak level of outcomes would occur with a smaller number of cashiers in A&TV as compared to WW. This argument is based on the greater customer service requirement in A&TV, and the higher frequency of sales transactions in WW. Our results supported this hypothesis for both customer satisfaction indices, but not for the number of sales transactions where the peak level was at the same point. This is surprising because we would have expected the longer average service times in A&TV to put a greater 'squeeze' on customer advice that is required before most purchases, with even a relatively small increase in the number of cashiers.

The second experiment investigated employee empowerment. During our case study work we observed the implementation of a new refund policy allowing cashiers to decide whether or not to make a refund up to the value of £50, rather than referring the authorization decision to a section manager. In our model we systematically varied the level of employee empowerment ranging from 0% (all refund decisions requiring a section manager) to a maximum 100% (all decisions made by the cashier), predicting higher levels of refund satisfaction and cashier utilization with greater levels of cashier empowerment. Our results for refund satisfaction were surprising in that there was a curvilinear relationship with empowerment, and also a difference between departmental peaks (50% for A&TV, 25% for WW). These results suggest that more complex interactions are occurring between our agents and some other constraining factors are occurring at the higher levels of empowerment. This may be linked to the empowered employees adhering to a stricter refund policy (resulting in less customer satisfaction). Cashier utilization increased with empowerment as predicted for A&TV, but interestingly this relationship was reversed in WW. This interesting but subtle pattern requires further investigation to determine whether or not it is statistically significant, or whether varying the level of empowerment does not result in a significant difference.

Our results so far are broadly consistent with our hypotheses, producing outcomes that we would expect based on our empirical observations of activity on the shop floor, and we are confident that our simulator in its current state produces valid results. Considering the abstract nature of the simulator we would not anticipate it to provide us with precise numerical figures regarding the performance of the simulated departments, but rather help us to understand the direction of the relationships between management practices and our outcome measures. This information can be linked to certain mechanisms and principles operating in the two departments that vary with different management practices (e.g. high to low empowerment). Secondary to understanding these management practices we are also able to draw comparisons between the operations of the departments due to their different configurations, which we attribute to inherent differences between product categories and the resulting work requirements.

## 5 ManPraSim v2: Enhancement of the existing simulator

### 5.1 Model implementation (ManPraSim v2)

We have verified ManPraSim v1 rigorously, in particular the in-built queuing system and the staff allocation and we have validated the simulator through the experiments described in the previous section. Based on this foundation we are now building a progressively more complex simulator (ManPraSim v2).

Our current development areas include:

- the addition of realistic footfall reflecting different patterns of customer footfall during the day and across different days of the week
- the addition of different kinds of customer (hereafter referred to as 'customer types')
- the introduction of a finite population of customers that can evolve over time
- the addition of internal and external stimuli that influence evolution and decision making

In this section we will discuss the first three enhancements which we have already implemented and are currently testing. The last enhancement is still work in progress and will be discussed in our next paper.

There are certain peak times where the pressure on staff members is higher, which puts them under higher work demands, and results in different service times. There is a weekly demand cycle. For example on a Saturday, a lot more customers visit the store compared to the average weekday. In our new model we have incorporated these real temporal fluctuations in customer arrival rates, across daily variations to opening hours. The model includes the calculated hourly footfall values for each of the four case study departments for each hour of the day and each day of the week, based on sales transaction data which are automatically recorded by the company. Conversion rates are based on staff estimates and data from a leading UK retail database. The gaps between customer arrivals, like in our original model, are based on exponential distributions which account for further variation in weekly footfall.

In ManPraSim v1 our agents are quite homogeneous. For each department we only use one set of distributions to describe the behavior of all customers of that particular department. In real life customers display certain shopping behaviors which can be categorized. Hence we enhance the realism of our agents' behavior by introducing customer types, which results in a heterogeneous customer base, thereby allowing us to test customer populations acting in a way closer to what we would find in reality. Customer types have been introduced based on the three the company uses for their analysis (shopping enthusiasts, solution demanders, service seekers) and have been expanded by the addition of two further types (disinterested shoppers, and internet shoppers who are customers that only seek advice but are likely to buy only from the cheapest website on the internet). The three company types have been identified by the case study organization as the customers who make biggest contribution to their business, in terms of both value and frequency of sales. In order to avoid over-inflating the amount of sales we have introduced the two additional types which use services but do not generally make so many purchases. The definition of each type is based on the customer's likelihood to perform a certain action, classified as either: low, moderate, or high. The definitions can be found in Table 1.

A moderate likelihood is equivalent to an average probability value in ManPraSim v1, and it is a threshold value for executing a decision-making process. These figures were based on staff estimates. The low and high likelihood thresholds are logically derived on the basis of this value (i.e. a new mode is calculated if the customer type's likelihood to execute a particular decision is not moderate). The same method is used for adapting delays which are defined by triangular frequency distributions.

In ManPraSim v1 our customer agents have only a limited lifespan. Once they leave the department the will be removed from the system. A key aspect to consider is that the most interesting system outcomes evolve over time and many of the goals of the retail company (e.g., service standards) are planned strategically over the long-term. We have therefore introduced a finite population where each agent is given a certain characteristic based on one out of five possible types mentioned above. Once agents are created they are added to a customer pool. Each hour a certain amount of agents chosen at random from the agents in the customer pool are released into the department based on the footfall definitions. Once customers are inside the shop it is business as usual. When they have finished shopping, statistics will be updated and the shopper returns to the customer pool. Only if customers have previously bought something can they go for refund. A customer retains his or her satisfaction index throughout the runtime of the simulation. To implement these new concepts we have added two new states to our customer state chart, one following the initialization of the customer agents to allow modeling a customer pool where all potential customers are gathered and one replacing the exit transition to model customers leaving the department. The transition between the two states converts a leaving customer into a potential customer that might enter the department again after a resting period.

Tab. 1 Definitions for each type of customer

| Customer type | Likelihood to | | | |
| --- | --- | --- | --- | --- |
| | buy | wait | ask for help | ask for refund |
| Shopping enthusiast | high | moderate | moderate | low |
| Solution demander | high | low | low | low |
| Service seeker | moderate | high | high | low |
| Disinterested shopper | low | low | low | high |
| Internet shopper | low | high | high | low |

We have also added some transitions that allow emulating the behavior of customers when the store is closing. These are immediate exits of a state that are triggered when the shop is about to close. Not all states have these additional transitions as it is for example very unlikely that a customer will leave the store immediately when he/she is already queuing to pay. Now the simulated department empties within a ten to fifteen minute period, which conforms to what we observed in the real system.

The staff agent logic has not been changed during the enhancement process.

### 5.2 Experiments (ManPraSim v2)

Equipped with a working version of our enhanced simulator ManPraSim v2 we have repeated some of the previous experiments to validate our new tool and we have added some new experiments to investigate the impact of different customer agent types.

To test the new version of our simulator we repeated the first experiment described in Section 4.2 to see how ManPraSim v2 behaves compared to ManPraSim v1. In this experiment we vary the number of cashiers whilst keeping the overall staffing level constant, using 'number of transactions', 'number of satisfied customers' and 'overall satisfaction level' as performance measures.

In general we would expect the number of transactions to be very similar between v1 and v2 of ManPraSim because we have tried to mimic the generic customer from v1 by using an even mix of all five customer types available in v2. With regards to the number of satisfied, neutral and unsatisfied customers we would expect the results to show a shift from neutral to either satisfied or unsatisfied. This is hypothesized to occur as ManPraSim v2 enables the customer population to re-enter the system and each re-entry will increase the likelihood that neutral customers will shift to satisfied or unsatisfied, leaving a diminishing pool of neutral customers as the number of re-entries increases. Looking at overall customer satisfaction we would expect similar trends for v1 and v2 of the simulator, but the magnitude of the results for v2 will be significantly higher because it incorporates an accumulated history of satisfaction trends for customers who have returned to the department on multiple occasions, unlike v1 which just records satisfaction levels for single, independent visits.

The numerical results for the experiments (for ManPraSim v1 and v2) are shown in Table 2 and a graphical representation of the results is presented in Figure 4.

Looking at the number of transactions for both departments, it is clear that both model versions produce a highly similar pattern of results. The number of satisfied customers is higher across all conditions of both departments in the later model version. This is as predicted and interestingly very high levels of satisfaction can be seen in WW in particular. We attribute this to the higher transaction volumes in WW coupled with our expectations of ManPraSim v2 resulting in higher levels of customer satisfaction as customers visit the store on multiple occasions and commit to polarized opinions. Examining the overall satisfaction level, our hypotheses hold; results for both departments clearly follow the same trends regardless of model version. In summary, all results are as predicted.

It is noteworthy that the real transaction data that we have received from the case study departments are higher than the ones we have seen in the results of our first experiment with the new simulator. This may be explained by the distribution of customer types. The departmental managers report that they find mainly shopping enthusiasts in the WW department while the A&TV department is mainly visited by solution demanders and service seekers. We have used this information in our second experiment. For this experiment the staffing consists of four cashiers and six normal staff for A&TV and five cashiers and five normal staff for WW (these settings showed the best transaction results in the previous experiment). We would assume that by using these stereotypes we get a value closer to the value of the real system. For A&TV we are not able to predict if it will be above or below the value of the real system as we would need to know the exact proportions of the two customer types. For the experiment we assumed a fifty-fifty split. For WW we would assume the number of transactions to be higher than the ones of the real system as we have created an entire population of shopping enthusiast with a high likelihood to buy, while in the real system there will be small groups of the other types present as well.

Tab. 2 Results for experiment 1

| Department | Staffing | | ManPraSim v1 | | | ManPraSim v2 | | |
|---|---|---|---|---|---|---|---|---|
| | cashiers | normal staff | transactions | number of satisfied customers | overall satisfaction level | transactions | number of satisfied customers | overall satisfaction level |
| A&TV | 1 | 9 | 4842 | 12360 | 9680 | 6020 | 13208 | -158043 |
| | 2 | 8 | 9885 | 14762 | 20257 | 11126 | 21752 | 67849 |
| | 3 | 7 | 14268 | 17408 | 28292 | 14191 | 25912 | 226339 |
| | 4 | 6 | 14636 | 17221 | 33200 | 14909 | 27187 | 314089 |
| | 5 | 5 | 13848 | 16123 | 28200 | 14475 | 24409 | 230997 |
| | 6 | 4 | 12890 | 14840 | 18472 | 13744 | 21817 | 99942 |
| | 7 | 3 | 11903 | 13500 | 8641 | 12631 | 17762 | -109958 |
| WW | 1 | 9 | 8096 | 18417 | 17070 | 10245 | 27356 | -219142 |
| | 2 | 8 | 15910 | 22639 | 42176 | 19492 | 46505 | 627058 |
| | 3 | 7 | 25346 | 28736 | 59098 | 25953 | 60225 | 1377537 |
| | 4 | 6 | 30448 | 32289 | 74320 | 28224 | 63247 | 1831069 |
| | 5 | 5 | 28758 | 30332 | 76830 | 28889 | 60698 | 1864292 |
| | 6 | 4 | 27399 | 28843 | 69021 | 27651 | 59024 | 1751009 |
| | 7 | 3 | 25652 | 26906 | 53526 | 26091 | 52173 | 1337444 |

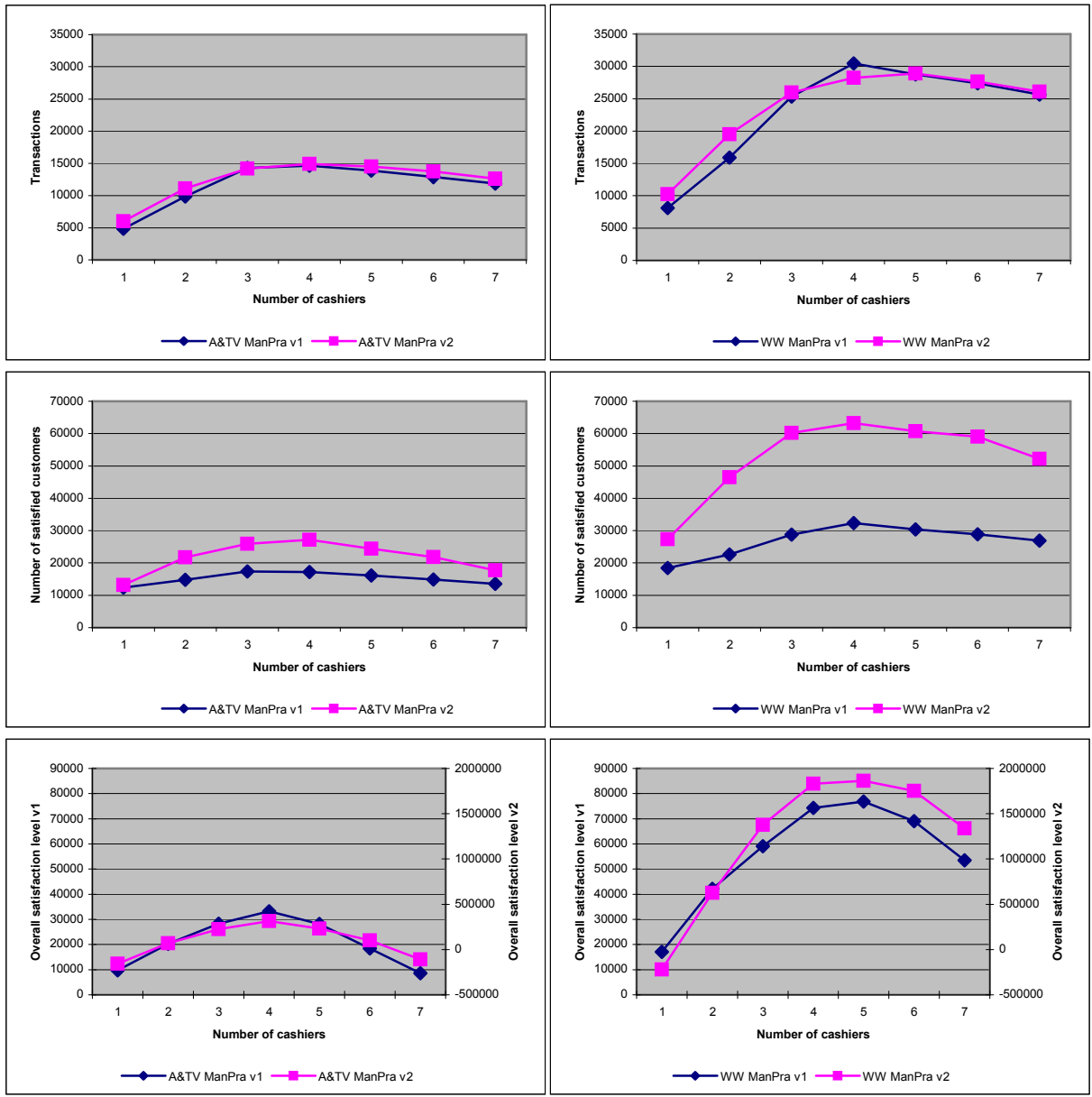

Fig. 4 Diagrams for experiment 1

The results for the experiment are shown in Table 3. They indicate that when we choose the specific types mentioned by the department managers to define our population (while keeping all other values constant) we get a significant change in the number of transactions. For both departments we get our hypothesized results although we are further away from the true value in the WW than predicted. It might be to extreme to assume a complete population of shopping enthusiasts.

Further experiments have to be conducted in order to clarify the impact of customer types on the perception of different management practices. Another aspect we have not yet implemented but we assume would have a big impact on transactions is flexible staffing and flexible till manning by cashiers. Currently, our allocation of staff is created in the beginning of a simulation run and the numbers do not change during the simulation run. But as we have introduced variable customer arrival rates particularly on Saturdays there is a much bigger demand for service staff and cashiers which in the real system is considered in the rota planning. This means that the real department can cope better with the additional service demand at peak times while the simulation cannot respond and will lose a lot of sales during peak times due to under staffing.

Tab. 3 Results for experiment 2 (using transaction of the real systems as 100% benchmark)

| Department | Customer types | Transactions Experiment 1 | Transactions Experiment 2 |
|---|---|---|---|
| A&TV | 50% solution demanders, 50% service seekers | 81.50% | 96.80% |
| WW | 100% shopping enthusiasts | 92.10% | 120.19% |

The new developments in our model open up at important question about the way in which we are measuring satisfaction, and whether or not this is desirable. Indeed, it makes sense that as customers re-visit a department they accumulate more information which helps them commit to an opinion, but the large effect we are seeing is also likely to be an artefact of the way in which this variable is measured. An individual customer's satisfaction is measured in whole numbers, ranging from negative whole numbers ('dissatisfaction') through zero ('don't know') and up through positive whole numbers ('satisfaction'). Therefore this measure is particularly sensitive to differences around the zero point on the scale: a one-point shift in either direction will change the label given to that customer's level of satisfaction. In future we plan to address this measurement issue and change the way in which individual customer satisfaction values are classified. By evaluating the overall distribution of scores from a large data set, we can assess the distribution of scores and benchmark customer satisfaction against equally-sized categories broken at more evenly spaced intervals, resulting in a more useful assessment of this outcome measure. For example, examining the distribution may result in a categorization such as: below -5 = very dissatisfied; -5 to -2 = dissatisfied; -1 to 1 = neutral; 1 to 5 = satisfied; above 5 = very satisfied.

The calibration of our simulator to validate it against the real data of the case study department turns out to be a difficult as we rely on many estimates and in addition it is difficult to pinpoint how much time a staff member allocates to each different tasks. Brooks and Shi [26] propose that if historical output data is available (in our case transaction data), one could also use the inverse method, such as using the output of a simulation to calibrate the inputs by comparing them to the output data of the real system. The problem with this method is that there are usually many solutions and no method of identifying the correct one. Furthermore, there might be measurement errors contaminating the historical data. The problem is that we have a large amount of input data (even more in our latest simulator) that we could tweak to receive the desired outputs for our simulator but we would not get an insight into the real operation of the department if we do not use our case study input data. What we have to do is a sensitivity analysis and then focus on finding more accurate estimates for the most sensible data. Preliminary tests have shown that conversion rates play a key role for the calibration.

Once our simulator has been calibrated we believe that our extended version of the simulator reflects the operation and the behavior of customers and staff within the departments in an abstract but realistic way and the results we gain by running virtual scenarios will be much closer to what we would expect to happen in the real system.

# 6 Conclusions

In this paper we present the conceptual design, implementation and operation of a department store simulator used to understand the impact of management practices on retail productivity. As far as we are aware this is the first time researchers have applied an agent-based approach to simulate management practices such as training and empowerment. Although our simulator draws upon specific case studies as the source of information, we believe that the general model could be adapted to other retail companies and other areas of management practices that involve significant human interaction.

After successfully building and testing a first version of our department store simulator (details can be found in [25,27]) we have now built a second version of the simulator. The new version includes more complex operational features to make it a more realistic representation closer to the real retail environment that we have observed. We also developed our agents with the goal of enhancing their intelligence and heterogeneity. To meet these goals we have introduced schedules, customer types and some simple form of evolution. The next step is to continue creating new scenarios to test with our simulator to enhance our understanding about the impact of management practices on retail productivity.

There is also some more development work planned. This includes the introduction of internal and external stimuli that influence evolution and decision making (as mentioned above) which allows our finite customer base to change their shopping preferences during runtime based on the service they receive and the service that others receive (word of mouth). Furthermore we plan to introduce costs, for example for additional equipment, training, wages for different staff levels and so on. We also have to review our service level index values and further intertwine our empirical case study work with academic theories from the marketing area.

The multi-disciplinary of our team has helped us to gain new insights into the behavior of organizations. In our view, the main benefit from adopting this approach is the improved understanding of and debate about a problem domain. The very nature of the methods involved forces researchers to be explicit about the rules underlying behavior and to think in new ways about them. As a result, we have brought work psychology and agent-based modeling and simulation closer together to form a new and exciting research area.